%% file: Example.tex
\documentclass[a4paper,twoside]{article}
\pdfoutput=1
\usepackage{amsmath}
\usepackage{bm}

\usepackage{booktabs}
\usepackage{float} 
\usepackage{placeins}
\usepackage{stfloats}

\usepackage{epsfig}
\usepackage{subcaption}
\usepackage{calc}
\usepackage{amssymb}
\usepackage{amstext}
\usepackage{amsmath}
\usepackage{amsthm}
\usepackage{multicol}
\usepackage{pslatex}
\usepackage{apalike}
\usepackage{algorithm2e}
\usepackage[bottom]{footmisc}
\usepackage{SCITEPRESS}     

\begin{document}

\title{Sparse Binary Representation Learning for Knowledge Tracing}

\author{\authorname{Yahya Badran\sup{1,2}\orcidAuthor{0009-0006-9098-5799}, Christine Preisach\sup{1,2}\orcidAuthor{0009-0009-1385-0585}}
\affiliation{\sup{1}University of Applied Sciences, Moltekstr. 30, 76133 Karlsruhe, Germany }
\affiliation{\sup{2}Karlsruhe University of Education, Bismarckstr 10,76133 Karlsruhe, Germany}
\email{\{yahya.badran, christine.preisach\}@h-ka.de}
}

\abstract{\input{sections/abstract}}
\onecolumn \maketitle \normalsize \setcounter{footnote}{0} \vfill

\section{\uppercase{Introduction}}
\label{sec:introduction}
\input{sections/intro}

\input{sections/related}

\input{sections/background}

\section{\uppercase{Model Description}}
\label{sec:model}
\input{sections/model}

\input{sections/experiments}

\section{\uppercase{Conclusions}}
\label{sec:conclusion}
\input{sections/conclusion}

\section*{\uppercase{Acknowledgements}}
We would like to thank the anonymous reviewers for their helpful feedback and suggestions. This work was funded by the federal state of Baden-Württemberg as part of the Doctoral Certificate Programme "Wissensmedien" (grant number BW6{\_}10).

\bibliographystyle{apalike}
{\small
\bibliography{example}}

\end{document}

%% file: sections/abstract.tex
Knowledge tracing (KT) models aim to predict students' future performance based on their historical interactions. Most existing KT models rely exclusively on human-defined knowledge concepts (KCs) associated with exercises. As a result, the effectiveness of these models is highly dependent on the quality and completeness of the predefined KCs. Human errors in labeling and the cost of covering all potential underlying KCs can limit model performance.

In this paper, we propose a KT model, Sparse Binary Representation KT (SBRKT), that generates new KC labels, referred to as auxiliary KCs, which can augment the predefined KCs to address the limitations of relying solely on human-defined KCs. These are learned through a binary vector representation, where each bit indicates the presence (one) or absence (zero) of an auxiliary KC. The resulting discrete representation allows these auxiliary KCs to be utilized in training any KT model that incorporates KCs. Unlike pre-trained dense embeddings, which are limited to models designed to accept such vectors, our discrete representations are compatible with both classical models, such as Bayesian Knowledge Tracing (BKT), and modern deep learning approaches.

To generate this discrete representation, SBRKT employs a binarization method that learns a sparse representation, fully trainable via stochastic gradient descent. Additionally, SBRKT incorporates a recurrent neural network (RNN) to capture temporal dynamics and predict future student responses by effectively combining the auxiliary and predefined KCs. Experimental results demonstrate that SBRKT outperforms the tested baselines on several datasets and achieves competitive performance on others. Furthermore, incorporating the learned auxiliary KCs consistently enhances the performance of BKT across all tested datasets.

%% file: sections/intro.tex
Knowledge tracing is a fundamental task in educational data mining that involves modeling a student's mastery of knowledge concepts (KCs) over time to predict future performance. Accurate knowledge tracing enables personalized learning experiences, adaptive tutoring systems, and informed instructional decisions. Traditional models, such as Bayesian Knowledge Tracing (BKT)\cite{bkt} and Deep Knowledge Tracing (DKT)\cite{dkt}, often rely on predefined KCs associated with each exercise. However, these predefined KCs may not fully capture the underlying complexities and latent structures of the learning process.

Recent advancements in representation learning have introduced the possibility of uncovering latent knowledge through data-driven methods by learning a new representation of the data. This approach has been utilized in different areas of machine learning including education data mining \cite{general_representation,pre_cont_embed2020}. By leveraging exercise embeddings and neural networks, we can identify hidden patterns and relationships that are not immediately apparent from predefined KCs alone. These representations can be further utilized in downstream tasks, which, in our case, involve using the learned representations to improve the performance of simpler models, such as BKT. 

In this paper, we propose a model that learns a representation that can be further utilized by other KT models. Specifically, we train a neural network to learn sparse binary vector for each exercise. These vectors can be used to extract further latent labels, which we refer to as auxiliary knowledge concepts (auxiliary KCs). In this context, a value of one in the binary vector indicates the presence of an auxiliary KC, while a value of zero indicates its absence. This can help mitigate lack of coverage in the human pre-defined KCs. This can help mitigate the possible lack of coverage in the human pre-defined KCs.

Each auxiliary KC clusters the coursework questions into two groups: those associated with the KC and those without. However, these auxiliary KCs do not possess explicit human labels. Instead, in this work, we use these auxiliary KCs in downstream tasks to improve the performance of other knowledge tracing (KT) models. By integrating these auxiliary KCs into models like BKT that assume independence between KCs, we aim to enhance their predictive capabilities without compromising their interpretability. Future work can investigate the possible advantages of displaying these KCs to students, as well as the potential to assign meaningful human labels to them with the help of large language models (LLMs) or expert teachers.



Our approach involves the following steps:

1. \textit{Knowledge tracing that learns a discrete and sparse representation}: We develop a neural network model that learns a sparse binary vector for each exercise based on student interaction data.

2. \textit{Extracting Auxiliary KCs from the binary representation}: The learned binary representation represent further latent discrete features that we call auxiliary KCs.

3. \textit{Integration with downstream models}: We utilize these auxiliary KCs in downstream knowledge tracing models, such as BKT and DKT to enhance their performance.

We evaluate our model on real-world educational datasets to assess its effectiveness in improving predictive performance over traditional methods. The results demonstrate that incorporating auxiliary KCs learned through our proposed model generally enhances the accuracy of knowledge tracing models, often yielding significant improvements.

The contributions of this paper are threefold:
\begin{itemize}
\item \textbf{Model Development}: We introduce a method for learning sparse binary representations, which can be used to extract auxiliary KCs. 

\item \textbf{Enhancement of downstream tasks}: We demonstrate how these learned representations can be utilized in downstream models like BKT, improving their predictive performance while maintaining simplicity and interpretability.

\item \textbf{Empirical Validation}: We perform extensive experiments on real-world datasets to demonstrate the effectiveness of our approach.
\end{itemize}

In the following sections, we review related work in knowledge tracing and representation learning, detail our methodology for learning and integrating auxiliary KCs, present experimental results, and discuss the implications of our findings for educational data mining.

%% file: sections/related.tex
\section{RELATED WORK}


Early approaches to knowledge tracing, such as Bayesian Knowledge Tracing (BKT) \cite{bkt}, utilize a Hidden Markov Model (HMM) to model each knowledge component (KC) independently. BKT represents student knowledge using two binary states: mastered and not mastered, with transitions between these states governed by interpretable probabilities for slip, guess, and learning. While this simple structure ensures high interpretability, BKT's strong independence assumption ignores dependencies between KCs, limiting its expressiveness. Furthermore, its straightforward probability-based framework often underperforms compared to deep learning-based models, which excel at capturing complex and intricate relationships between KCs \cite{whendeep,Howdeep}.


Deep Knowledge Tracing (DKT) \cite{dkt} introduced a neural network-based approach to knowledge tracing by employing a recurrent neural networks (RNNs) to model temporal knowledge state change during coursework. DKT was followed by numerous models that utilized different deep learning architectures such as transformers like architecture \cite{akt} and memory based neural networks \cite{dkvmn}.

Dynamic Key-Value Memory Networks (DKVMN) \cite{dkvmn} uses a static key memory, a fixed matrix, which remain consistent across interactions to capture the inherent relationships between exercises and underlying concepts. Alongside this, a dynamic value memory is maintained to track the evolving mastery levels of these concepts as students engage with exercises. DKVMN does not utilize the human predefined KCs in the dataset, instead they consider the static key memory to be a model of the latent concepts in the dataset.


Given that KCs are human-predefined tags and therefore prone to errors, some studies aim to mitigate potential inaccuracies by learning methods to correct or calibrate these human-defined KCs \cite{calib_q_matrix,calib_q_matrix2,calib_q_matrix3}. This approach differs from our work, as we do not attempt to modify or correct the contributions of human experts. Instead, we focus on augmenting the existing KCs with newly derived auxiliary KCs to enhance model performance.

Papers such as \cite{pre_cont_embed2020,cont_emb2} propose models that learn vector embeddings for downstream knowledge tracing tasks. However, these embeddings are dense, making them less interpretable and limiting their applicability to deep learning models. In contrast, our approach learns discrete, sparse binary representations that map each question to a new set of auxiliary KCs. These representations are versatile, as they can be utilized by both deep learning and classical models, where the latter can offer greater interpretability.

In \cite{end2end_binary}, the authors propose a method to learn new KCs for each question, fully replacing the original KCs defined by experts. This approach aligns with methods that aim to recalibrate human-defined KCs. Each question is represented as a binary vector, where a value of one indicates that the corresponding dimension is a KC associated with the question. However, their model does not directly produce a binary representation. Instead, it relies on a continuous representation with a regularization term to encourage closeness to a binary vector, with binarization applied only after training. In contrast, our model explicitly learns a binary representation to define new auxiliary KCs. While our approach could be adapted to entirely replace the original KCs, this is not the primary objective of our research.

%% file: sections/background.tex
\section{BACKGROUND}

In this section, we provide an overview of knowledge tracing models that are relevant to this work. 
\input{sections/bkt}
\input{sections/dkt}

%% file: sections/bkt.tex
\subsection{Bayesian Knowledge Tracing}

Bayesian Knowledge Tracing (BKT)\cite{bkt} is formulated as a Hidden Markov Model (HMM), where the learner's knowledge state is represented as a latent variable. This probabilistic framework predicts whether a learner has mastered a given KC based on their observed responses to practice opportunities.

BKT models the learning process as an HMM with two states as follows:
\begin{itemize}
    \item \textbf{Knowledge State (Latent):} Whether the learner has mastered the KC ($K_t = 1$) or not ($K_t = 0$).
    \item \textbf{Observed Performance:} Correct ($O_t = 1$) or incorrect ($O_t = 0$) response at time $t$.
\end{itemize}

For a detailed derivation of the BKT model, see \cite{bkt_intro,bkt}. The model can be defined using the following HMM parameters:
\begin{itemize}
    \item $P(L_0)$: The prior probability of mastery before any practice.
    \item $P(T)$: The learning probability, i.e., the chance of transitioning from non-mastery ($K_t = 0$) to mastery ($K_{t+1} = 1$) after a practice opportunity.
    \item $P(G)$: The guess probability, i.e., the likelihood of a correct response given non-mastery.
    \item $P(S)$: The slip probability, i.e., the likelihood of an incorrect response given mastery.
\end{itemize}

The key HMM transitions are as follows:
\begin{itemize}
    \item Transition from non-mastery to mastery is governed by $P(T)$.
    \item Mastery is assumed to be \textbf{absorbing}: once a learner masters the KC, they remain in mastery indefinitely.
\end{itemize}

BKT assumes that KCs are independent. That is, the knowledge or mastery of one KC does not influence another. This assumption simplifies the model but may not reflect realistic learning scenarios where skills often interrelate. Moreover, BKT does not model forgetting behavior, the the likelihood of transitioning from mastery to non-mastery is zero under this model.

\subsubsection{Forgetting Variant}

A common extension to BKT incorporates forgetting, where the model allows transitions from mastery back to non-mastery. Different approaches has been used to incorporate forgetting to BKT \cite{BKTvariants,pybkt}. As described in \cite{Howdeep}, one approach is to add a new forgetting probability, $P(F)$, which models the probability of transitioning from mastery to non-mastery as follows:

\begin{equation}
P(K_{t+1} = 0 \mid K_t = 1) = P(F)
\end{equation}
This variant accounts for scenarios where learned knowledge decays over time. In this work, we exclusively use this variant, and the term BKT will refer to this specific version.


%% file: sections/dkt.tex
\subsection{Deep Knowledge Tracing}

Deep Knowledge Tracing (DKT) \cite{dkt} employs a recurrent neural network (RNN), typically a Long Short-Term Memory (LSTM) network, to process sequences of learner interactions and predict future performance. Each input sequence consists of interaction pairs \((q_t, y_t)\), where $q_t$ is a single KC in the original DKT and $y_t$ is the correctness of the learner's response (\(y_t = 1\) for correct, \(y_t = 0\) for incorrect).

The RNN processes these sequences to update a hidden state \(h_t\), which represents the learner's evolving knowledge state. The hidden state is updated iteratively:
\begin{equation}
h_t = f(h_{t-1}, x_t),
\end{equation}
where \(x_t = (q_t, y_t)\) encodes the interaction at time \(t\), and \(f\) represents the RNN dynamics (e.g., LSTM or GRU).

Using the updated hidden state \(h_t\), the model predicts the probability of a correct response for each KC:
\begin{equation}
P(y_{t+1} = 1 | q_{t+1}, h_t)
\end{equation}
This prediction is generated through a fully connected output layer followed by a softmax activation.

While DKT generally outperforms classical models such as BKT \cite{Howdeep,whendeep} and makes fewer assumptions about the data, it has notable limitations. One significant drawback is its lack of interpretability. Unlike BKT, which provides clear parameter interpretations, such as learning and slip probabilities, the hidden state in DKT is opaque, making it difficult to extract meaningful insights.

Lastly, the original model implementation can suffer from label leakage and requires specialized methods to be correctly evaluated \cite{pykt}. To avoid this issue, in this paper, $q_t$ represents the set of KCs associated with a given question, and $x_t$ is defined as the mean of the embeddings of these KCs. In this paper, DKT denotes this variant.

Deep Knowledge Tracing marks a paradigm shift in knowledge tracing, offering a flexible and powerful approach to modeling student learning. However, challenges such as interpretability and reliance on large datasets maintain the relevance of classical models like BKT. In this paper, we aim to bridge the gap by transferring the representational strengths of deep learning to enhance classical models such as BKT.

%% file: sections/model.tex
\begin{figure}[!h]
\centering
{\epsfig{file = 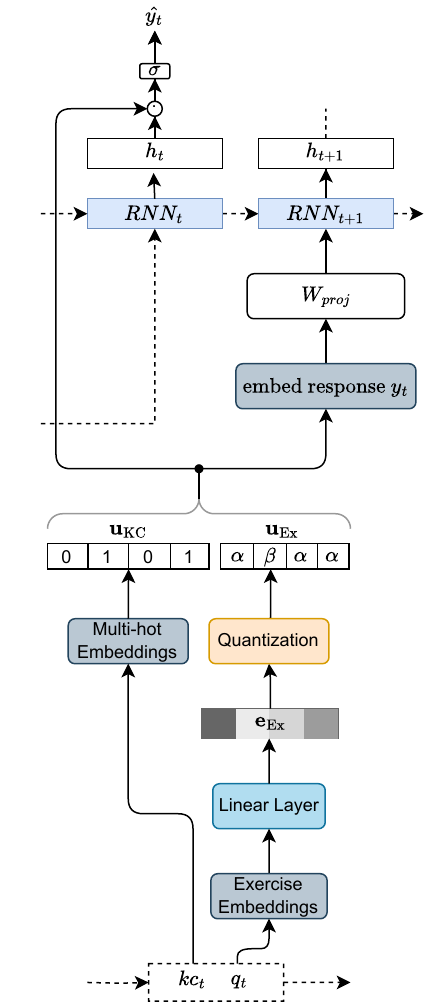, width =0.7\linewidth}}
\caption{The overall architecture of the proposed model.}
\label{fig:main}
\end{figure}

In this section, we present our model for predicting student responses. The model integrates KCs and exercise embeddings to capture the temporal dynamics of student interactions. Central to our approach is the learning of a binary representation, where each bit represents the presence or absence of a KC. This binary representation serves as a multi-hot encoding, a vector in which multiple positions can be "hot" (i.e., set to one) to indicate the presence of multiple KCs simultaneously.

The key components of the model include the construction of these binary multi-hot vectors, derived from KCs and exercise embeddings, a quantization algorithm for generating the binary vectors, and a prediction mechanism using an RNN. The overall architecture is described in Figure \ref{fig:main}.

\subsection{Multi-Hot Encoding of Knowledge Concepts}

Let $N$ be the total number of KCs in the dataset. For each exercise, we construct an $N$-dimensional multi-hot vector $\mathbf{u}_{\text{kc}} \in \{0, 1\}^{N}$ to represent its associated KCs. Additionally, we construct a vector $\mathbf{u}_{(\text{kc},y)}$ that embeds the correctness label \( y \in \{0, 1\} \), where \( y = 1 \) corresponds to a correct response and \( y = 0 \) corresponds to an incorrect response.
The multi-hot vector \( \mathbf{u}_{\text{kc}} \) is defined as:

\begin{equation}
\mathbf{u}_{\text{kc}}[i] =
\begin{cases} 
1, & \parbox[t]{0.6\linewidth}{\raggedright if the $i$-th knowledge concept is associated with the exercise,} \\
0, & \text{otherwise.}
\end{cases}
\end{equation}

Using \( \mathbf{u}_{\text{kc}} \), we construct the final multi-hot vector \( \mathbf{u}_{(\text{kc},y)} \) as:

\begin{equation}
\mathbf{u}_{(\text{kc},y)} = \mathbf{u}_{\text{kc}} \cdot y \oplus \mathbf{u}_{\text{kc}} \cdot (1 - y),
\end{equation}

where \( \oplus \) denotes vector concatenation. This formulation ensures that the first \( N \) entries of \( \mathbf{u}_{(\text{kc},y)} \) encode the KCs for a correct response (\( y = 1 \)), while the last \( N \) entries encode the KCs for an incorrect response (\( y = 0 \)).

\subsection{Multi-Hot Encoding from Exercise Embeddings}

Our goal is to map each exercise $q$ to a binary vector, similar to the KC-based $\mathbf{u}_{kc}$, where a value of one indicates the presence of an auxiliary KC and a value of zero indicates its absence.

\subsubsection{Exercise Embeddings and Linear Layer}

Each exercise is represented by an embedding vector $\mathbf{x}_{Ex} \in \mathbb{R}^d$, where $d$ is the dimensionality of the embedding space. The embedding $\mathbf{x}_{Ex}$ is passed through a linear layer to obtain a latent representation:

\begin{equation}
\label{eq:linear}
\mathbf{e_{Ex}} = \mathbf{W} \mathbf{x}_q + \mathbf{b},
\end{equation}

where $\mathbf{W} \in \mathbb{R}^{M \times d}$ is a weight matrix, $\mathbf{b} \in \mathbb{R}^M$ is a bias vector and $M$ is the number of auxiliary KCs. The output $\mathbf{e_{Ex}} \in \mathbb{R}^M$ captures the latent features of the exercise.

\subsubsection{The Quantization Algorithm}
\label{sec:algo}

The latent vector $\mathbf{e_{Ex}}$ is transformed into a discrete vector $\mathbf{u}_{Ex} \in \{\alpha, \beta\}^M$, where the model learns the scalar values of $\alpha$ and $\beta$ under the constraint $\alpha > \beta$. In downstream tasks, $\alpha$ represents one, indicating the presence of an auxiliary KC, and $\beta$ represents zero, indicating its absence. Additionally, the representation is constrained to be sparse, ensuring that only a few ones appear in the final vector. These ones correspond to auxiliary KCs in downstream tasks.

\textbf{Algorithm Steps:}

1. \textbf{Top-$C_{\text{max}}$ Selection:} Identify the indices of the top $C_{\text{max}}$ largest values in $\mathbf{e_{Ex}}$, denoted as $\mathcal{I}_{\text{top}} \subseteq \{1, 2, \dots, M\}$. Create a mask $\mathbf{m} \in \{0, 1\}^M$ where:
\begin{equation}
   \mathbf{m}[i] =
   \begin{cases}
   1, & \text{if } i \in \mathcal{I}_{\text{top}}, \\
   0, & \text{otherwise}.
   \end{cases}
\end{equation}

   This mask will enforce the sparsity constraint in the next step.

2. \textbf{Binary Transformation:}
Rather than using traditional binary neural network methods such as the sign function, our method outputs two possible values at each dimension, denoted by $\alpha$ and $\beta$  with the constraint that $\alpha > \beta$. To achieve this, we apply the following discretization function $Q$ to $\mathbf{e}_{Ex}$:

\begin{equation}
Q(\mathbf{e}_{Ex}) = f(\mathbf{e}_{Ex}) \cdot \mathbf{m}
\end{equation}

Here, $f$ is applied element-wise to each dimension of $\mathbf{e}_{Ex}$ and is defined as:

\begin{equation}
f(x) =
\begin{cases} 
0, & \text{if } x \leq 0, \\
1, & \text{if } x > 0.
\end{cases}
\end{equation}

Lastly, we apply the following:

$$\mathbf{u}_{Ex} = 
Q(e_{Ex})\alpha  + (1-Q(e_{Ex}))\beta
$$
where \( \beta = c \sigma(p_{\beta}) \) and \( \alpha = c (1 + \sigma(p_{\alpha})) \). Here, \( c \) is a hyperparameter (set to one in our implementation). \( p_{\alpha} \) and \( p_{\beta} \) are scalar trainable parameters used to generate \( \alpha \) and \( \beta \), respectively.

\textbf{Training with Gradient Descent:}

To enable training of this discrete mapping using stochastic gradient descent, we apply the \textbf{straight-through estimator (STE)} \cite{ste}. In the forward pass, the quantization algorithm is applied as described in \ref{sec:algo}. In the backward pass, we modify the gradient by treating the discretization function $Q$ as the identity function:
\begin{equation}
\frac{\partial \mathbf{Q(e_{Ex})}}{\partial \mathbf{e_{Ex}}} = \mathbf{I},
\end{equation}
allowing gradients to flow through the discrete operation.

\subsubsection{Embedding Ground-Truth Labels}

Similar to how we embed ground truth labels with the multi-hot KC vector $\mathbf{u}_{(KC,y)}$, we construct the vector $\mathbf{u}_{(Ex,y)}$ to incorporate the correctness label $y$ by concatenating as follows:
\begin{equation}
\mathbf{u}_{(Ex,y)} = \mathbf{u}_{Ex} \cdot y \oplus \mathbf{u}_{Ex} \cdot (1 - y)
\end{equation}

\subsection{Sequence Modeling}  

At each time step \(t\), we concatenate the labeled multi-hot vectors \(\mathbf{u}_{(KC, y)}\) and \(\mathbf{u}_{(Ex, y)}\) to form a unified feature vector \(\mathbf{v}_t\) of dimension \(2N + 2M\):

\begin{equation}
\mathbf{v}_t = \mathbf{u}_{(KC,y)} \oplus \mathbf{u}_{(Ex,y)} 
\end{equation}


The concatenated vector $\mathbf{v}_{t}$ is projected into a dense representation suitable for sequence modeling through a linear transformation:
\begin{equation}
\mathbf{z}_{t} = \mathbf{W}_{\text{proj}} \mathbf{v}_{t},
\end{equation}
where $\mathbf{W}_{\text{proj}} \in \mathbb{R}^{D \times (2N + 2M)}$ is a trainable weight matrix, and $\mathbf{z}_{t} \in \mathbb{R}^D$ is the resulting dense feature vector.

To model temporal dependencies, we use an RNN, which offers a lower computational cost compared to other sequential models, such as Transformers. The sequence of dense feature vectors \( \{\mathbf{z}_{1}, \mathbf{z}_{2}, \dots, \mathbf{z}_{T}\} \) is fed into a recurrent neural network (RNN):

\begin{equation}
\mathbf{h}_{t} = \text{RNN}(\mathbf{z}_{t}, \mathbf{h}_{t-1}),
\end{equation}

where \( \mathbf{h}_{t} \) represents the hidden state at time step \( t \).

The hidden state \( \mathbf{h}_{t} \) is subsequently processed through a linear transformation to produce logits corresponding to each KC and auxiliary KC:

\begin{equation}
\mathbf{o}_{t} = \mathbf{W}_{\text{out}} \mathbf{h}_{t} + \mathbf{b}_{\text{out}},
\end{equation}

where \( \mathbf{W}_{\text{out}} \in \mathbb{R}^{(N + M) \times H} \) is the weight matrix, and \( \mathbf{b}_{\text{out}} \in \mathbb{R}^{N + M} \) is the bias vector, both of which are trainable parameters.

To compute the final prediction, we concatenate the binary multi-hot vectors \( \mathbf{u}_{\text{KC}, t} \in \{0, 1\}^N \) and \( \mathbf{u}_{\text{Ex}, t} \in \{0, 1\}^M \):

\begin{equation}
\mathbf{u}_{t} = \mathbf{u}_{\text{KC}, t} \oplus \mathbf{u}_{\text{Ex}, t} \in \{0, 1\}^{N + M},
\end{equation}

where \( \oplus \) denotes the concatenation operation. The predicted probability of a correct response is then calculated as:

\begin{equation}
\hat{y}_{t} = \sigma\left( \mathbf{u}_{t}^\top \mathbf{o}_{t} \right),
\end{equation}

with \( \sigma(\cdot) \) denoting the sigmoid activation function.

\subsection{Application of the Learned Binary Representation in Downstream Tasks}

The discrete vector $\mathbf{u}_{\text{Ex}}$ can serve as features for downstream tasks. As it only contains two values $\alpha$ and $\beta$, it can be easily mapped to ones and zeros. Given that our model is constrained with $\alpha > \beta$, we simply map $\alpha$ to one which denotes an existence of an auxiliary KC and $\beta$ to zero to denote the lack of an auxiliary KC.

%% file: sections/experiments.tex
\section{EXPERIMENTS}
In this section, we evaluate the effectiveness of our proposed model. We compare its performance with baseline models and demonstrate the utility of the extracted auxiliary KCs in downstream tasks. All experiments were conducted on publicly available educational datasets.

\subsection{Experimental Setup}

\subsubsection{Datasets}

We use widely used and publicly available datasets for knowledge tracing:
\begin{itemize}
    \item \textbf{ASSISTments2009}\footnote{\url{https://sites.google.com/site/assistmentsdata/home/}}:  It is derived from the the ASSISTments online learning platform which was gathered during the school year 2009-2010. They provide two datasets, the one we used is called the skill-builder data.
    \item \textbf{ASSISTments2017}\footnote{\url{https://sites.google.com/view/assistmentsdatamining/}}: It is a much recent data from ASSISTments and it was used for the Workshop on Scientific Findings from the ASSISTments Longitudinal Data Competition during the The 11th Conference of Educational Data Mining. However, we utilized the publicly available preprocessed version used in \cite{akt}.
    \item \textbf{Algebra2005} \cite{algebra05}: This dataset was part of the 2010 KDD Cup Educational Data Mining Challenge.
    \item \textbf{riiid2020} \cite{riiid}: It was introduced as part of a Kaggle competition aimed at improving AI-driven student performance prediction. The dataset consists of millions of anonymized student interactions with an AI-based tutoring system, focusing on question-solving activities. We choose a million entry from this dataset.
\end{itemize}

Detailed statistics can be found in table~\ref{table:stats}

\begin{table}[H]
\caption{Dataset attributes after prepossessing.}
\label{table:stats}
\centering
\begin{tabular}{llllll}
\toprule
\textbf{dataset} & questions & KCs  & students \\ \midrule
ASSISTments2009         & 17751           & 123           & 4163               \\
ASSISTments2017         & 3162           & 102           & 1709               \\
Riiid2020       & 13522          & 188          & 3822                        \\ 
Algebra2005        & 173650          & 112           & 574                                  \\
\end{tabular}
\end{table}

\subsubsection{Baselines}

We compare our model against the following baseline methods:
\begin{itemize}
    \item \textbf{Bayesian Knowledge Tracing (BKT)}\cite{bkt} : A probabilistic model that uses predefined KCs for student modeling.
    \item \textbf{Deep Knowledge Tracing  with average embeddings} : A neural network-based model that learns directly from student response sequences without requiring predefined KCs. To avoid accounting for label leakage we simply average the KC embeddings for all KCs belonging to the same question.
    \item \textbf{Dynamic Key-Value Memory Networks (DKVMN)}\cite{dkvmn}:  employs a memory network structure with two types of memory: key memory, representing latent knowledge concepts, and value memory
    \item \textbf{Deep Item Response Theoy (deepIRT)}\cite{deepirt}: It incorporate a model similar to DKVMN with Item Response Theory (IRT) which is a psychometric approach that models the relationship between student abilities, question difficulty, and the probability of answering correctly
    \item \textbf{Question-centric interpretable KT model (QIKT)}\cite{qikt}: Which is another model that combines IRT with deep learning.
\end{itemize}

\subsubsection{Implementation Details}

For our proposed model, we use an embedding size of $d = 32$ for both dense and binary exercise embeddings, which means we have $32$ auxiliary KCs. We used the Long short-term memory (LSTM) type of RNN architecture with a hidden size of $h = 128$.  We use a maximum of $C_{\text{max}} = 4$ auxiliary KCs per exercise. We train the models using the Adam optimizer with a learning rate of $0.001$ for all models except for BKT which we used  $0.01$ (BKT was trained using stochastic gradient descent). We used a batch size of $32$ for DKVMN, deepIRT, and QIKT. We used a batch size of $128$ for DKT and BKT. All experiments were conducted using a dataset split of 80\% for training, 10\% for validation, and 10\% for testing. We use the area under the curve (AUC) metric for evaluation.

\subsection{Results}

\subsubsection{Performance Comparison}

Table \ref{table:pre} summarizes the performance of our proposed model in comparison to the baseline methods. The results indicate that our model consistently outperforms all baselines on certain datasets, while achieving the second-best performance on the remaining datasets. These findings underscore the efficacy of our approach, demonstrating that despite the discrete constraints, our model can surpass others that depend on dense representations.


\input{tables/pretrain_results}

\subsubsection{Downstream Task Performance with BKT}

\input{tables/downstream_table}

We assess the utility of the extracted auxiliary KCs by training BKT and DKT models on these representations, as shown in Table \ref{table:downstream}. The results reveal that BKT augmented with auxiliary KCs (BKT+aux) outperforms the standard DKT on both the assistment2009 and riiid2020 datasets. Furthermore, the inclusion of auxiliary KCs boosts BKT's performance across all datasets, although the improvement on the algebra2005 dataset is minimal. In contrast, DKT with auxiliary KCs (DKT+aux) achieves better performance on all datasets except algebra2005, where it underperforms compared to the original DKT.


\input{sections/ablation}

%% file: tables/pretrain_results.tex
\begin{table*}[!ht]
\centering
\caption{AUC Scores with Highlighted and Marked Winners and Runners-Up (* Best, ** Second Best).}
\label{table:pre}
\begin{tabular}{lllll}
\toprule
Model & Algebra2005 & ASSISTment2009 & ASSISTment2017 & riiid2020 \\
\midrule
BKT & 0.7634 & 0.6923 & 0.6081 & 0.6215 \\
DKT & 0.8198 & 0.7099 & 0.6807 & 0.6503 \\
DKVMN & 0.7759 & 0.7362 & 0.7169 & \textit{0.7362}** \\
DeepIRT & 0.7750 & 0.7374 & 0.7170 & 0.7360 \\
SBRKT & \textit{0.8223}** & \textbf{0.7602}* & \textit{0.7494}** & \textbf{0.7369}* \\
QIKT & \textbf{0.8335}* & \textit{0.7574}** & \textbf{0.7527}* & 0.7324 \\
\bottomrule
\end{tabular}
\end{table*}

%% file: tables/downstream_table.tex
\begin{table*}[bp]
\caption{AUC Scores with Winners and Runners-Up Highlighted (* Best, ** Second Best). DKT+aux and BKT+aux refer to DKT and BKT models augmented with pretrained auxiliary KCs.}
\label{table:downstream}
\centering
\begin{tabular}{lllll}
\toprule
Model & Algebra2005 & ASSISTment2009 & ASSISTment2017 & riiid2020 \\
\midrule
BKT & 0.7634 & 0.6923 & 0.6081 & 0.6215 \\
BKT+aux & 0.7655 & \textit{0.7325}** & 0.6760 & \textit{0.7173}** \\
DKT & \textbf{0.8198}* & 0.7099 & \textit{0.6807}** & 0.6503 \\
DKT+aux & \textit{0.7997}** & \textbf{0.7481}* & \textbf{0.7422}* & \textbf{0.7365}* \\
\bottomrule
\end{tabular}
\end{table*}

%% file: sections/ablation.tex
\subsection{Ablation Study}
To demonstrate the effectiveness of the quantization layer, we create three variants:

\begin{itemize}
    \item \textbf{SBRKTtanh}: This variant adds the hyperbolic tangent function (tanh) as an activation function to the linear transformation in \ref{eq:linear}. The discretization step maps the output to either -1 or +1 (instead of $\alpha$ and $\beta$).
    \item \textbf{SBRKT10}: This variant applies a sigmoid activation to the output of the linear transformation in \ref{eq:linear}. Discretization is performed by mapping any value less than 0.5 to 0, and values greater than or equal to 0.5 to 1 (instead of $\alpha$ and $\beta$).
    \item \textbf{SBRKTdense}: This variant completely removes the quantization step, utilizing a dense representation instead. However, this representation cannot be directly used in downstream tasks with the approach outlined in this paper.
\end{itemize}

As shown in Table \ref{table:pre_ablation}, the proposed model outperformed all other variants, except on the algebra2005 dataset, where it achieved the second-best performance with a small difference (0.006). Notably, QCKTdense significantly underperformed compared to the other models, highlighting the importance of the quantization step in this architecture.

\input{tables/ablation_pretrain}

We also carried out experiments to evaluate the utility of these variant representations in downstream tasks. As shown in Table \ref{table:downstream_ablation}, the models that utilize the auxiliary KCs of SBRKT outperformed all other variants, except in the algebra2005 data set, where none of the variants showed significant improvement, and some even experienced performance drop when auxiliary KCs were added.

\input{tables/ablation_downstream}

\subsection{Summary of Findings}

Our experiments reveal the following key insights:
\begin{itemize}
    \item The proposed model outperforms the baselines across multiple benchmarks, despite the sparse discrete constraints imposed by the architecture which helped extract further auxiliary KCs.
    \item The extracted auxiliary KCs can significantly enhance downstream tasks. In our experiments, BKT consistently showed improved performance when the learned auxiliary KCs were incorporated across all datasets. However, the same was not true for DKT. While DKT+aux achieved substantial performance gains on some datasets (e.g., an increase of over 6\% in AUC on the Assistment2017 dataset), it underperformed on the algebra2005 dataset.
\end{itemize}

%% file: tables/ablation_pretrain.tex
\begin{table}[H]
\caption{AUC Scores with Highlighted and Marked Winners and Runners-Up (* Best, ** Second Best). algebra05, assist09, and assist17 correspond to Algebra2005, ASSISTments2009, and Assistments2017, respectively. SBR, SBR10, and SBRtanh represent SBRKT, SBRKT10, and SBRKTtanh, respectively.}
\label{table:pre_ablation}
\begin{tabular}{llll}
\toprule
Dataset & algebra2005 & assist09 & assist17 \\
\midrule
SBR & \textit{0.8223}** & \textbf{0.7602}* & \textbf{0.7494}* \\
SBR10 & \textbf{0.8231}* & \textit{0.7464}** & 0.7431 \\
SBRdense & 0.8122 & 0.7169 & 0.7448 \\
SBRtanh & 0.8166 & 0.7449 & \textit{0.7491}** \\
\bottomrule
\end{tabular}
\end{table}

%% file: tables/ablation_downstream.tex
\begin{table}[H]
\caption{AUC Scores with highlighted and marked winners and runners-up (* Best, ** Second Best). algebra05, assist09, and assist17 correspond to Algebra2005, ASSISTments2009, and Assistments2017, respectively. +AX10, +AXtanh means trained with auxiliary KCs from SBRKT10 and SBRKTtanh, respectively.}
\label{table:downstream_ablation}
\begin{tabular}{llll}
\toprule
Dataset & algebra05 & assist09 & assist17 \\
\midrule
BKT & 0.7634 & 0.6923 & 0.6081 \\
DKT & \textbf{0.8198}* & 0.7099 & 0.6807 \\
BKT+aux & 0.7655 & \textit{0.7325}** & 0.6760 \\
DKT+aux & \textit{0.7997}** & \textbf{0.7481}* & \textbf{0.7422}* \\
BKT+AX10 & 0.7745 & 0.7283 & 0.6577 \\
DKT+AX10 & 0.7899 & 0.7318 & \textit{0.7301}** \\
BKT+AXtanh & 0.6860 & 0.6736 & 0.5969 \\
DKT+AXtanh & 0.7454 & 0.6974 & 0.6722 \\
\bottomrule
\end{tabular}
\end{table}

%% file: sections/conclusion.tex
In this paper, we proposed Sparse Binary Representation Knowledge Tracing (SBRKT), a model designed to overcome the limitations of predefined KCs by generating auxiliary KCs through a learnable sparse binary representation. Our results demonstrated that SBRKT achieves competitive performance across datasets and significantly enhances downstream models' capabilities.

In particular, integrating the auxiliary KCs with Bayesian Knowledge Tracing (BKT) outperformed the original Deep Knowledge Tracing (DKT) model on some datasets, while preserving the interpretability, which is a hallmark of BKT. The auxiliary KCs also improved BKT's predictive performance across all tested datasets without requiring modifications to its foundational structure.


Our findings underscore the potential of SBRKT to address limitations in traditional KT models while expanding their applicability in real-world educational scenarios. Future research could explore refining auxiliary KC generation methods and leveraging these representations in broader applications for adaptive learning and personalized education systems.

